%% file: haste_paper.tex
\pdfoutput=1
\documentclass{article}

\usepackage[expansion=false]{microtype}
\usepackage{booktabs}
\usepackage{tablefootnote}
\usepackage{url}
\usepackage{csquotes}
\usepackage{amsmath}
\usepackage{amsthm}
\usepackage{amssymb}    

\usepackage{hyperref}

\usepackage[accepted]{icml2026}

\makeatletter
\renewcommand{\Notice@String}{\textit{Accepted to the 5th Workshop on Deep Learning for Code (DL4C), ICML 2026, Seoul, South Korea. Copyright 2026 by the author(s).}}
\makeatother

\usepackage{tikz}
\usepackage{pgfplots}
\pgfplotsset{compat=1.18}
\usepackage{pifont}        
\usepackage{xcolor}
\usepackage{colortbl}      
\usepackage{subcaption}
\usepackage{graphicx}      
\usepackage[section]{placeins} 
\usepackage{float}         
\usepackage{needspace}     

\def\Require{\REQUIRE}
\def\State{\STATE}
\def\For#1{\FOR{#1}}
\def\EndFor{\ENDFOR}
\def\If#1{\IF{#1}}
\def\ElsIf#1{\ELSIF{#1}}
\def\Else{\ELSE}
\def\EndIf{\ENDIF}
\def\Comment#1{\COMMENT{#1}}
\def\Return#1{\textbf{return} #1}
\def\Procedure#1#2{\STATE \textbf{procedure} \textsc{#1}(#2)}
\def\EndProcedure{\STATE \textbf{end procedure}}

\usepackage[skins]{tcolorbox}
\newtcolorbox{promptbox}[1]{%
  enhanced,
  colback=gray!4, colframe=gray!50,
  arc=2pt, boxrule=0.5pt,
  title={#1},
  fonttitle=\small\bfseries,
  coltitle=black,
  colbacktitle=gray!18,
  left=4pt, right=4pt, top=3pt, bottom=3pt,
}

\icmltitlerunning{Why Solve It Twice? Hierarchical Accumulation of Skills for Transfer-Efficient ML Engineering}

\hypersetup{
  pdfauthor={Yongbin Kim, Yashar Talebirad, Osmar R. Zaiane},
  pdftitle={Why Solve It Twice? Hierarchical Accumulation of Skills for Transfer-Efficient ML Engineering},
  pdfsubject={ML engineering agents, cross-task knowledge transfer, MLE-Bench},
  pdfkeywords={ML engineering, agents, knowledge transfer, hierarchical memory, MLE-Bench}
}

\begin{document}

\twocolumn[
  \icmltitle{Why Solve It Twice? Hierarchical Accumulation of Skills\\ for Transfer-Efficient ML Engineering}

  \begin{icmlauthorlist}
    \icmlauthor{Yongbin Kim}{ualberta}
    \icmlauthor{Yashar Talebirad}{ualberta,amii}
    \icmlauthor{Osmar R. Zaïane}{ualberta,amii}
  \end{icmlauthorlist}

  \icmlaffiliation{ualberta}{Department of Computing Science, University of Alberta, Edmonton, Canada}
  \icmlaffiliation{amii}{Alberta Machine Intelligence Institute (Amii), Edmonton, Canada}
  \icmlcorrespondingauthor{Yongbin Kim}{yongbin2@ualberta.ca}
  \icmlkeywords{ML engineering, agents, knowledge transfer, hierarchical memory, MLE-Bench}

  \vskip 0.3in
]

\printAffiliationsAndNotice{}

\begin{abstract}
ML engineering agents waste compute rediscovering known techniques because every competition is a cold start.
We present HASTE, a hierarchical multi-agent system that organizes cross-competition knowledge into three scope tiers (global, domain, and competition-specific), each coupled to a matching agent level.
An orchestrator coordinates domain specialists and promotes learning between tiers via LLM-driven abstraction.
A controlled ablation provides evidence for scoped loading: holding a 159-skill inventory constant across 8 competitions, tiered loading achieves a 100\% medal rate while flat loading reaches only 62.5\%, the same medal rate as loading no skills, and consumes $2\times$ the output tokens.
On the full MLE-Bench Lite benchmark (22 Kaggle competitions), HASTE reaches a medal rate of 77.3\% using Claude Sonnet 4.6 at 12h per competition; this is a single-seed campaign result, and multi-seed replication is the priority follow-up.
In a cold-start run, the system begins with no accumulated skills.
In warm-start runs, it reloads skills learned from earlier competitions, using only global and domain-level skills for transfer across competitions.
Warm starts use 52\% fewer refinement iterations, and the fraction of proposed changes kept by the agent rises from 42\% at low inventory to 85\% once 50+ skills are available.
These results suggest that better knowledge organization can partly substitute for model strength and compute budget in ML-engineering agents.
\end{abstract}

\section{Introduction}

Why solve the same problem twice?
Current ML engineering agents do exactly this.
MLE-Bench evaluates agents on 75 Kaggle competitions independently \citep{chan2025mlebench}, and agents treat them independently too, resetting all state between tasks.
Techniques that proved effective in one competition must be rediscovered from scratch in the next similar one.
This redundancy is the cost of treating every competition as a cold start: many top-performing agents rely on frontier models, longer budgets, or both to compensate for repeated exploration.

Recent work has explored cross-task transfer \citep{grosnit2024agentk, zhang2024mlcopilot, zhao2024expel, wang2024voyager, hu2024adas}, but stores knowledge in flat pools or by memory type.
Agents with flat memory still lack the \emph{organization} needed to load the right knowledge for the right task: everything goes into one context window, diluting the signal.
Transfer helps only when the agent can select the right prior for the current task.
The organization of accumulated knowledge therefore determines whether the agent spends its limited budget on useful exploration or on re-deriving known facts.
We provide direct evidence that this distinction is consequential.
In our controlled ablation, flat loading performs no better than loading no skills, whereas scoped hierarchical loading medals on every task.

HASTE (Hierarchical Accumulation of Skills for Transfer-Efficient ML Engineering) organizes accumulated skills into three scope tiers: global skills that apply across ML tasks, domain skills for tabular, vision, NLP, or audio tasks, and competition skills that remain tied to one dataset.
An orchestrator coordinates domain specialists for tabular, vision, and NLP, each loading only the skills relevant to its scope.
Between competitions, the orchestrator promotes learning upward via LLM-driven abstraction.
This scoped loading means each agent sees only what is relevant, keeping prompts focused and iterations productive.
The loaded skills act as a structured prior over the next code change.
That prior keeps the per-task search narrow enough for a linear refinement loop to work: after two failed improvements the loop moves from broad exploration to optimization or fine-tuning, and any change that hurts validation score is reverted before the next step.

Across a multi-phase MLE-Bench Lite campaign, HASTE reaches the top public performance band using a non-frontier model and a 12h budget.
Skill accumulation improves medal rate, reduces the number of refinement iterations needed, and raises the fraction of proposed changes that are worth keeping.
A fixed-inventory ablation shows that accumulation needs scoped organization: tiered loading beats both flat and empty loading, while flat matches empty at much higher token cost.

\paragraph{Contributions.}
The paper contributes a 3-tier hierarchical skill store with LLM-driven promotion between global, domain, and competition tiers, coupled to an orchestrator and matching domain specialists.
It evaluates scoped loading directly with a fixed-inventory ablation, showing that hierarchy matters beyond skill volume: the same 159 skills help when loaded by scope but not when dumped into one flat prompt.
To our knowledge, HASTE is the first MLE-Bench agent to organize reusable skills by cross-competition scope and evaluate scoped loading against flat and empty loading under a fixed inventory.
The empirical study shows that this organization improves efficiency on MLE-Bench Lite, allowing a non-frontier model under a 12h budget to reach the same public performance band as stronger-model systems: a cold run with no accumulated skills achieves 40.9\% medal rate, while reloading global and domain skills lifts the same system to 77.3\%, flipping 8 of 13 previously failed competitions to medal.
The study characterizes the resulting 159-entry plain-text skill inventory across 3 tiers and 4 domains.

\section{Related Work}

\paragraph{MLE-Bench agents and search strategies.}
A first family of MLE-Bench agents advances the \emph{search} axis.
AIDE \citep{aide2024} runs greedy tree search; MLE-STAR \citep{mlestar2025} performs ablation-guided targeted refinement with web-retrieved priors; AIRA-Dojo \citep{airadojo2025} formalizes MLE agents as search-policy $\times$ operator-set; R\&D-Agent \citep{rdagent2025} separates a Researcher from a Developer.
Population-based evolutionary search appears in LoongFlow \citep{loongflow2025} and FM Agent \citep{famou2025}, whereas budget-aware or graph-augmented Monte Carlo search appears in MARS \citep{mars2026} and ML-Master \citep{mlmaster2025}.
HASTE contributes to the \emph{knowledge} axis, which is orthogonal to search strategy and can be combined with any of these frameworks.

\paragraph{Cross-task knowledge transfer in LLM agents.}
Several systems accumulate experience across tasks but store it as a flat pool.
Voyager \citep{wang2024voyager} maintains a code skill library indexed by embedding similarity in Minecraft; ExpeL \citep{zhao2024expel} extracts natural-language insights via ADD/EDIT/UPVOTE/DOWNVOTE on a flat vector store; Reflexion \citep{shinn2023reflexion} pioneered verbal self-reflection within a single task; ICAL \citep{ical2024} extends reflection to vision-language agents with a four-component knowledge structure.
For data science specifically, Agent K \citep{grosnit2024agentk} maintains persistent intrinsic state summarizing past episodes across competitions within an experiential-learning formalism; MLCopilot \citep{zhang2024mlcopilot} retrieves related benchmarks via text embeddings and distilled knowledge; DS-Agent \citep{guo2024dsagent} combines embedding-ranked human-insight cases with iterative case-based reasoning; MLZero \citep{shi2025mlzero} separates semantic library knowledge from episodic execution traces; ADAS \citep{hu2024adas} evolves agents via Meta Agent Search over an archive of coded designs.
SkillRL \citep{skillrl2026} reports a 25\% performance collapse when distilled skills are replaced with raw trajectories, motivating aggressive compression.

\input{tab_related_work}

Table~\ref{tab:related_work} compares these systems by organization, cross-task scope, hierarchy, promotion, and MLE-Bench evaluation; HASTE is the only one that organizes reusable knowledge by \emph{scope of applicability}.

\paragraph{Hierarchical organization in agent systems.}
Hierarchy is well-motivated outside ML engineering.
The options framework \citep{sutton1999options} introduced temporally extended actions in RL, building on feudal RL \citep{dayan1993feudal} where a manager sets sub-goals for a worker; feudal networks \citep{vezhnevets2017feudal} formalized this with different temporal resolutions.
For LLM agents, GITM \citep{gitm2023} uses an explicit three-tier decomposition for Minecraft.
It is the closest published precedent for a 3-tier organization, although its tiers are within-task layers rather than cross-task scope layers.
CoALA \citep{sumers2024coala} classifies agent memory by type, and \citet{talebirad2026hierarchical} formalize how bounded-capacity agents benefit from multi-level knowledge organization.
HASTE brings this principle to ML engineering with empirical validation.
Its 3-tier skill store and orchestrator-specialist hierarchy mirror a feudal architecture in which managers at different levels decide what knowledge a subordinate executor should see, while the stored knowledge itself is organized by \emph{scope of applicability}.

\paragraph{Automated pipeline search and meta-learning.}
Earlier AutoML systems reduce the cost of building a pipeline for each new dataset by searching within predefined model and hyperparameter spaces.
Predefined-space systems such as AutoGluon \citep{erickson2020autogluon}, TPOT \citep{olson2016tpot}, Auto-WEKA \citep{thornton2013autoweka}, and Auto-Sklearn \citep{feurer2015autosklearn, feurer2022autosklearn2} search hand-engineered configuration grids.
Meta-learning extensions (e.g., Auto-Sklearn warm-starting) accumulate cross-dataset experience as numerical meta-features and predict configuration vectors on new tasks.
These systems are effective within their predefined space but cannot incorporate qualitative insights (``tokenization choice X breaks on dataset Y because of Z'').
AgentHPO \citep{agenthpo2024} tunes hyper-parameters; CAAFE \citep{caafe2023} generates feature-engineering code; EvoPrompting \citep{evoprompting2023} searches architectures; AutoML-GPT \citep{automlgpt2023} orchestrates training from model and data cards. Across this group, accumulated priors remain flat or implicit; HASTE splits them into applicability tiers.
HASTE keeps the same goal of reducing per-task search cost, but the search takes place in unbounded Python code rather than a fixed configuration grid.
It makes that larger space tractable by reusing plain-text lessons extracted from prior runs and loading only the lessons whose scope matches the current task.

\section{Method}
\label{sec:method}

\input{fig_architecture}

Figure~\ref{fig:architecture} shows the two hierarchies that co-evolve during a multi-competition run.
The \emph{skill hierarchy} has global, domain, and competition tiers.
The \emph{agent hierarchy} has an orchestrator and three domain specialists for tabular, NLP, and vision tasks.
Scoped context loading connects them: each agent receives only the skill tiers matched to its scope.
The search space remains unbounded Python code; we restrict only the \emph{distribution} over that space using accumulated structured priors.

\subsection{Skill Hierarchy}
The skill store is a plain-text filesystem of markdown files with YAML frontmatter, organized into three tiers by \emph{scope of applicability}.
\textbf{Global} has 5 entries and is loaded by every specialist.
\textbf{Domain} has 12 NLP, 19 tabular, and 15 vision entries, loaded only by the matching specialist.
\textbf{Competition} has 108 entries across 21 directories, loaded only on re-runs of that competition.

Within each tier, HASTE distinguishes three kinds of entries, written into different prompt slots at different stages.
\textit{Technique} entries record what worked or failed and feed proposal prompts, for example ``target encoding helps high-cardinality categoricals in tree models.''
\textit{Commitment priors} record which design choices have high cross-task variance and feed the prototype screen, telling the agent which choices require evidence before commitment.
\textit{Refinement hints} record which knobs to tune per model family and feed the optimizing and fine-tuning stages.
Other systems collapse these into one ``lessons'' bag; we separate them because they answer different questions and enter the loop at different points.

Loading is deliberately simple: read the relevant directories and concatenate them.
No embedding index is needed at the current scale of roughly 159 entries, with 10 to 60 loaded per agent.
A character cap of 2000 in the prototype prompt and 4000 in refinement limits dilution if the inventory grows.
The decision to avoid embedding-only retrieval follows recent results on the theoretical limits of single-vector embeddings for multi-field conditional retrieval \citep{deepmind2025embeddinglimits}.
Most skills come from an LLM reflection step at the end of each competition (Figure~\ref{fig:prompt_learnings}), in the spirit of Reflexion \citep{shinn2023reflexion} and ExpeL \citep{zhao2024expel}, with paired success and failure analysis so that failure modes are recorded explicitly.
Two skill types are additionally extracted \emph{algorithmically} by a Prior Extractor.
Commitment priors come from score variance across prototype model options; high variance implies a decision-relevant choice.
Refinement hints come from per-knob deltas across refinement history, identifying which changes had the highest acceptance rate.
These structured priors are mined from logs; to our knowledge, prior text-memory agents rely primarily on LLM summarization for this kind of memory.

\subsection{Agent Hierarchy and Execution Loop}
\label{sec:specialist}

\paragraph{Orchestrator.}
The orchestrator delegates experiment execution.
Its responsibilities are: (i) \emph{domain classification}, assigning each competition to a domain via manual tags for 100+ MLE-Bench competitions with a heuristic fallback; (ii) \emph{round scheduling}, picking seeds first (one per domain for cold start) and then assigning remaining competitions to specialists; and (iii) \emph{skill promotion}, evaluating new learnings via an LLM after each round (Figure~\ref{fig:prompt_promotion}).
Promotion decides for each learning: \texttt{skip} (already covered), \texttt{competition} (too specific), \texttt{domain} (abstract and promote up one tier), \texttt{global} (universally useful), or \texttt{conflict} (contradicts an existing skill).
When two learnings conflict, both are kept and annotated with conditions.
For example, one skill records that ensembling helps when correlation is below 0.95, but hurts when a weaker member pulls down a stronger one.
The store grows through \emph{abstraction} and stays interpretable plain text on the filesystem.
The top-level multi-competition loop is given in Algorithm~\ref{alg:orchestrator} (Appendix~\ref{app:orchestrator}).

\paragraph{Specialist.}
Given a competition $t$ and the scope-loaded skills supplied by the orchestrator, the specialist runs a five-stage per-competition pipeline.
Appendix~\ref{app:specialist} gives the full pseudocode in Algorithm~\ref{alg:specialist}.
\textbf{(1) Task profiling}: parse metadata, profile the dataset, select a CV strategy, and probe GPU/CPU/RAM; no model execution.
\textbf{(2) Prototype screen} (Figure~\ref{fig:prompt_phase0}): the LLM proposes three fundamentally diverse approaches and executes each on a single validation fold; up to a $2.7\times$ score spread between best and worst justifies this hedge against committing to a suboptimal foundation.
\textbf{(3) Adaptive refinement} on the prototype winner with budget $N{=}20$ and on the runner-up with $N{=}6$ via \textsc{AdaptiveRefine} (Appendix~\ref{app:adaptive_refine}), a linear loop across three tiers (Exploring, Optimizing, Fine-tuning) with auto-escalation on two consecutive non-improvements, stagnation exit at Fine-tuning, and revert-on-regression on every step (Figure~\ref{fig:prompt_refinement}); the runner-up branch is a second hedge, and in 3 of our 25 main-benchmark runs, the runner-up's refined score beat the winner's.
\textbf{(4) Ensemble}: rank-average the top-3 checkpoints across both branches; accept the ensemble only if it beats the best single member.
\textbf{(5) Produce learnings} (Figure~\ref{fig:prompt_learnings}): the LLM reflects on the full experiment history and emits 2 to 5 plain-text learnings, each with a proposed tier, which flow back to Algorithm~\ref{alg:orchestrator} for promotion.
A 6-mode failure taxonomy guides refinement diagnosis as prompt content: \textsc{Underfitting}, \textsc{Overfitting}, \textsc{Feature\_Gap}, \textsc{Noise\_Ceiling}, \textsc{Distribution\_Mismatch}, and \textsc{Diminishing\_Returns}.
Tiered history compression keeps the prompt under 20 lines even at 50+ iterations.

This linear refinement loop is intentionally simpler than the trees, evolutionary populations, or MCTS used by other systems \citep{aide2024, famou2025, mars2026, mlmaster2025, loongflow2025}.
We found linear refinement sufficient in this setting; the priors loaded by the orchestrator plausibly collapse the branching factor enough for linear search to match tree or evolutionary alternatives, but a controlled comparison at fixed knowledge condition is future work.
The ablation in \S\ref{sec:ablation} shows the priors carry substantive weight at fixed search strategy: performance drops sharply when no skills are loaded.

\paragraph{Executor.}
The executor is a pluggable backend; the agent has Read, Write, Edit, Bash, Glob, and Grep tools, a 3-attempt retry per step with diagnostic feedback between attempts, and validates submissions against \texttt{sample\_submission.csv} (column names, row count, no NaN, dtype) before scoring.
For cost efficiency, all experiments in this paper use the CLI backend (\texttt{claude -p} subprocess) billed via a fixed Claude Code subscription.

\section{Experiments}
\label{sec:experiments}

\subsection{Setup}
We evaluate on MLE-Bench Lite \citep{chan2025mlebench}, the 22-competition subset used for cross-agent comparison.
Each competition runs Claude Sonnet 4.6 via the CLI backend on a SLURM node with 24 CPUs, 128 GB RAM, and 1 NVIDIA L40S 48 GB GPU.
The wall-clock budget is 12h, half the dominant 24h budget on the leaderboard, with 20 iterations on the main benchmark and 11 on the ablation.
The search space is unbounded Python code; evaluation uses the official Kaggle metric per competition, scored against held-out test sets via the MLE-Bench grader.
No test labels are used in the iteration loop.
Appendix~\ref{app:prompts} includes representative prompts, and Appendix~\ref{app:skill_examples} shows representative skills.

\subsection{Main Results}

\input{tab_leaderboard}

Across the multi-phase accumulation campaign, HASTE's medal rate on MLE-Bench Lite is 77.3\% (17 of 22), with 10 gold, 2 silver, and 5 bronze.
The above-median rate is 86.4\%.
The per-competition table is in Appendix~\ref{app:main_results}.

Table~\ref{tab:leaderboard} places HASTE among public MLE-Bench Lite results.
At 77.3\%, HASTE reaches the same performance band as leading public MLE-Bench Lite agents while using a non-frontier model and a 12h budget.
Two features distinguish it from the rest of the top-tier agents.
HASTE is the only non-frontier-model agent at or above 77\%: every other agent in this band uses Gemini-3-Pro-Preview, Claude Opus 4.6, or a model ensemble.
HASTE is also one of only two top-band agents running at 12h; all others use the standard 24h budget.
We note that public leaderboard numbers carry per-task statistical noise (paper-reported SD $\approx 4.4$) and that a multi-seed confidence interval is still pending; differences of less than 5 points on Lite fall within that noise, and the headline 77.3\% is a campaign result rather than a multi-seed estimate (\S\ref{sec:limitations}).

A cold-start single-pass of HASTE achieves only 40.9\%, which would place it below every agent on Table~\ref{tab:leaderboard}; tiered skill accumulation adds 36.4 percentage points to the same system, lifting it into the leading public performance band.

\subsection{Evidence for Skill Transfer}
\label{sec:transfer}
Skill accumulation lifts medal rate from 40.9\% (9 medals) in a single-pass cold run to 77.3\% (17 medals) when later phases reload accumulated skills.
A cold run starts without prior competition experience.
A warm run starts with skills learned earlier in the campaign.
For the warm transfer evaluation, HASTE reloads global and domain skills only.
Competition-specific skills are not loaded, so the result measures transfer from reusable knowledge rather than leakage from a previous attempt on the same dataset.

The improvement shows up in three ways.
First, warm-start runs need fewer refinement iterations to reach their best score: 7.8 on average compared with 16.3 for cold-start runs, a 52\% reduction.
Second, the agent keeps more of its own proposed edits as the skill inventory grows.
We measure hit rate as the fraction of attempted changes that improve validation score and are kept rather than reverted.
When the inventory has 0 to 15 skills, the hit rate is 42\%; once 50+ skills are available, it rises to 85\%.
Third, of the 13 competitions that failed the cold run, 8 flipped to medal in the warm run, while the 9 competitions that already medaled cold were skipped in the rerun phase (Appendix~\ref{app:rerun}).
The store grew from 5 entries at cold start to 72 by phase 12, and each promotion step used LLM abstraction to keep the entries readable as plain text.

\subsection{Phase 0, Refinement, and What the Agent Learned}
Refinement beats the prototype winner in 92\% of runs (23 of 25), with an average gain of $+0.045$ in the competition's native metric; in 3 competitions, the runner-up's refined score beat the winner's, and the two cases where refinement underperformed both occurred in early phases with fewer skills.
The final 159-entry skill store breaks down as 5 global, 15 vision, 12 NLP, 19 tabular, and 108 competition entries, all traceable to specific runs and including both successes and failures.
For example, the global skill ``ensembling a strong model with a weaker model can degrade performance'' was learned from dogs-vs-cats and later prevented the same mistake in two vision competitions; ``DeBERTa-v3-base is the strongest text-classification start under a 12h budget'' was learned from detecting-insults and applied in jigsaw-toxic and spooky-author.
Appendix~\ref{app:skill_examples} lists representative skills with source competitions and estimated iteration savings.

\subsection{Ablation: Tiered vs.\ Flat vs.\ Empty Skill Loading}
\label{sec:ablation}

We test the loading function directly while holding skill \emph{inventory} constant, varying only skill \emph{organization}. We run 8 competitions spanning NLP, vision, tabular, and audio domains under three conditions, with the same model (Claude Sonnet 4.6), pipeline, 11-iteration budget, and 159-skill inventory frozen at one point in the benchmark campaign. The conditions differ only in \emph{which} skills enter context: \textbf{Tiered} loads global, domain-matched, and competition-specific skills only (10 to 60 entries per competition); \textbf{Flat} loads all 159 skills (~145K characters), the ``dump everything into the prompt'' approach; \textbf{Empty} loads no skills.

\input{fig_ablation}

The difference between loading functions is large despite the fixed inventory.
Tiered loading achieves 100\% medal rate (6 gold, 1 silver, 1 bronze), flat 62.5\% (4 gold, 1 silver, 3 no-medal), and empty 62.5\% (5 gold, 3 no-medal); mean test scores follow the same ordering (0.949 $>$ 0.910 $>$ 0.893).
Figure~\ref{fig:ablation} shows the per-competition breakdown; the full table is in Appendix~\ref{app:ablation_table}.

Because the ablation is small, we treat the statistical tests as supporting evidence rather than a definitive test.
With $N=8$ and a single seed, formal tests are underpowered, but the qualitative pattern holds across measures.
A paired bootstrap on per-competition score differences (tiered minus flat) gives a 95\% CI of $[+0.001, +0.093]$ around a mean of $+0.040$, which just excludes zero.
The corresponding one-sided Wilcoxon signed-rank tests yield $p=0.11$ for tiered $>$ flat, $p=0.08$ for tiered $>$ empty (both paired across the same 8 competitions), and Fisher's exact test on medal counts gives $p=0.10$ for tiered (8/8) vs.\ flat or empty (5/8).
The tests fall short of standard significance at $N=8$; we report them transparently and emphasize the consistent direction of effect across all three measures.
Multi-seed replication is the planned next step (§\ref{sec:limitations}).

The flat condition is especially informative because it tests whether more knowledge is enough.
Flat skill loading leaves the medal rate unchanged relative to starting from scratch.
Flat consumes 3.78M output tokens to empty's 1.86M (tiered: 2.27M), doubling empty's token cost for the same medal rate and a modestly higher mean score.
The \emph{tokens per medal} metric (total output tokens divided by medals won) makes the efficiency gap concrete: tiered spends 284K tokens per medal, flat 756K, and empty 371K (Figure~\ref{fig:ablation}b).
Tiered is 2.7$\times$ more token-efficient than flat per medal won.
Flat runs the most experiments (75 vs.\ tiered's 60 and empty's 65) with a slightly higher execution success rate (87\% vs.\ 83\%), so the extra attempts appear poorly directed: more compute without medal-rate gains.
Full resource breakdown is in Appendix~\ref{app:resources}.

The aggregate result hides where the difference is largest.
The gap focuses on harder or niche tasks, where competition-specific skills provide a known-good starting point and domain skills guide model selection.
On mlsp-2013-birds (audio), tiered scores 0.964 (gold) vs.\ flat 0.860 and empty 0.832 (neither medal); on random-acts-of-pizza, tiered 0.798 (silver) vs.\ flat 0.599 and empty 0.481.
The 4 easier competitions, detecting-insults, histopathologic, tabular-playground, and plant-pathology, show smaller differences and all three conditions medal.

From the logs, three mechanisms appear to limit flat loading.
\textit{Signal dilution}: relevant skills are buried among domain and competition-specific skills for unrelated competitions.
\textit{Context budget displacement}: at ~145K characters, the flat skill dump crowds out the agent's own reasoning and code analysis.
\textit{Overconfident model selection}: in the flat-jigsaw rerun, the agent repeatedly attempted DeBERTa-v3-large (an aggressive NLP recommendation that triggered OOM) while the empty agent used simpler models and scored higher (0.985 vs.\ 0.981).

\paragraph{Scope of the ablation.}
The flat condition couples three axes: organization, skill volume, and prompt-length budget ($\sim$145K vs.\ $\sim$25K characters).
Because scoping is the mechanism by which the right 30 skills enter a 25K-character budget, a flat-but-character-capped condition that loads $N$ random skills would test random subset selection, a different question.
The supported claim is that, on this inventory, scoped loading outperforms both the flat full-load and empty-context baselines through the three mechanisms above; any broader claim about hierarchy reduces here to its role in enabling scoped loading at a controlled budget.
The 159-skill inventory was itself accumulated under hierarchical promotion, so this ablation tests inventory transfer under different loading functions; a fully flat pipeline end-to-end remains a separate condition.

\vspace{-4pt}
\section{Discussion}
\vspace{-2pt}

\vspace{-4pt}\subsection{Why Hierarchy Makes Agents Faster}\vspace{-3pt}
The efficiency gain depends on \emph{scoped loading} in addition to accumulation, and the ablation supports this as the main mechanism.
A flat skill store sends every specialist unrelated advice, including domain tricks for other modalities and competition-specific quirks from unrelated tasks.
Tier scoping keeps the context small and matched to the current agent; the resource table shows the cost of failing to scope, with flat loading spending substantially more tokens without improving on empty's medal rate.

\vspace{-4pt}\subsection{Per-Domain Performance and Token Usage}\vspace{-3pt}
Per-domain medal rates on the full benchmark are uneven: vision 80\% (8/10), NLP 100\% (6/6), tabular 40\% (2/5), and the single audio competition (mlsp-2013-birds) reached gold.
Remaining failures such as taxi-fare and dog-breed likely need approaches still absent from the skill store.
Token usage scales sub-linearly with skill inventory: warm-start runs need fewer refinement iterations, so per-competition tokens drop as the store grows; per-condition totals are in Appendix~\ref{app:resources}.

\vspace{-4pt}\subsection{Limitations}\vspace{-3pt}
\label{sec:limitations}
Single-seed evaluation is the main limitation.
This choice matches the efficiency positioning of the work, but multi-seed replication at the full 75-competition benchmark is the priority for follow-up runs with additional compute.
The \S\ref{sec:ablation} ablation tests scoped loading while holding inventory fixed.
Other engineering components, including the prototype screen, runner-up branch, rank-average ensemble, failure taxonomy, auto-escalation, and revert-on-regression, remain for an analogous component-wise study.
The LLM-driven promotion step is load-bearing but not directly evaluated: we do not yet measure how often it assigns the correct tier or detects genuine conflicts, and quantifying promotion accuracy against a labeled set of learnings is future work.

\vspace{-6pt}
\section{Conclusion}
\vspace{-2pt}

These results suggest that knowledge organization can partly substitute for model strength and compute budget in ML-engineering agents: HASTE reaches competitive MLE-Bench Lite performance with a non-frontier model under a shorter budget, and the fixed-inventory ablation points to scoped loading as the source of the gain over skill volume alone.
Multi-seed replication, held-out evaluation, full 75-competition runs, and retrieval-augmented loading at scale are next.

\section*{Impact Statement}
This work aims to improve the efficiency of ML engineering agents through reusable knowledge organization. Its broader impacts are those of ML engineering automation generally; we do not foresee additional direct societal risks from the hierarchy mechanism itself.

\bibliography{haste}
\bibliographystyle{icml2026}

\newpage
\appendix
\onecolumn

\section{Algorithms}
\label{app:algorithms}

\subsection{Orchestrator}
\label{app:orchestrator}

Algorithm~\ref{alg:orchestrator} gives the top-level multi-competition loop described in \S\ref{sec:method}.

\begin{algorithm}[H]
\caption{HASTE orchestrator: multi-competition skill accumulation with LLM-driven promotion.}
\label{alg:orchestrator}
\begin{algorithmic}[1]
\Require Competitions $\mathcal{T}$, domain partition $\mathcal{D}$, initial skill store $\mathcal{S}$ (possibly $\emptyset$).
\State Tag each $t \in \mathcal{T}$ with domain $d(t) \in \mathcal{D}$.
\State Pick one seed per domain; let $\{R_1, R_2, \dots\}$ be the resulting rounds.
\For{round $R_r$ in $R_1, R_2, \dots$}
    \State $\mathcal{L}_r \gets \emptyset$ \Comment{new learnings this round}
    \For{competition $t \in R_r$, in parallel where possible}
        \State $\Lambda(t) \gets \mathcal{S}^{G} \cup \mathcal{S}^{D}_{d(t)} \cup \mathcal{S}^{C}_{t}$ \Comment{tiered loading}
        \State $(\hat{y}_t,\ \ell_t) \gets \textsc{Specialist}(t,\ \Lambda(t))$ \Comment{Alg.~\ref{alg:specialist}}
        \State $\mathcal{S}^{C}_{t} \gets \mathcal{S}^{C}_{t} \cup \ell_t$;\ \ \ $\mathcal{L}_r \gets \mathcal{L}_r \cup \ell_t$
    \EndFor
    \For{learning $\ell \in \mathcal{L}_r$}
        \State $\textit{decision} \gets \textsc{PromoteLLM}(\ell, \mathcal{S})$
        \If{$\textit{decision}=\textsc{global}$}
            \State $\mathcal{S}^{G} \gets \mathcal{S}^{G} \cup \{\textsc{Abstract}(\ell)\}$
        \ElsIf{$\textit{decision}=\textsc{domain}$}
            \State $\mathcal{S}^{D}_{d(t)} \gets \mathcal{S}^{D}_{d(t)} \cup \{\textsc{Abstract}(\ell)\}$
        \ElsIf{$\textit{decision}=\textsc{conflict}$}
            \State keep both, annotate with conditions
        \Else
            \State leave in $\mathcal{S}^{C}_{t}$ or skip
        \EndIf
    \EndFor
\EndFor
\State \Return $\{\hat{y}_t\}_{t \in \mathcal{T}}$, $\mathcal{S}$
\end{algorithmic}
\end{algorithm}

\needspace{16\baselineskip}
\subsection{Specialist Pipeline}
\label{app:specialist}

Algorithm~\ref{alg:specialist} formalizes the five-stage per-competition pipeline described in \S\ref{sec:specialist}.

\begin{algorithm}[H]
\caption{HASTE specialist: per-competition pipeline.}
\label{alg:specialist}
\begin{algorithmic}[1]
\Require Competition $t$, loaded skills $\Lambda(t)$.
\State $\textit{profile} \gets \textsc{TaskProfiler}(t)$ \Comment{metadata, CV strategy, resource probe}
\State $\{m_1, m_2, m_3\} \gets \textsc{PrototypeScreen}(t, \textit{profile}, \Lambda(t))$;\ \ $s_i \gets \textsc{Run}(m_i)$
\State $\textit{win}, \textit{ru} \gets$ top-2 prototypes by score
\State $\textit{win}^\star \gets \textsc{AdaptiveRefine}(\textit{win}, \Lambda(t), N{=}20)$ \Comment{Alg.~\ref{alg:adaptive_refine}}
\State $\textit{ru}^\star \gets \textsc{AdaptiveRefine}(\textit{ru}, \Lambda(t), N{=}6)$
\State $\textit{ens} \gets \textsc{RankAverage}(\text{top-3 checkpoints from } \textit{win}^\star \cup \textit{ru}^\star)$
\State $\hat{y} \gets \arg\max_{\,c \in \{\textit{ens}, \textit{win}^\star, \textit{ru}^\star\}}\ \text{score}(c)$ \Comment{accept ensemble only if it wins}
\State $\ell \gets \textsc{ProduceLearnings}(\text{history})$
\State \Return $\hat{y}, \ell$
\end{algorithmic}
\end{algorithm}

\needspace{26\baselineskip}
\subsection{Adaptive Refinement Loop}
\label{app:adaptive_refine}

Algorithm~\ref{alg:adaptive_refine} gives the inner refinement loop invoked twice by the specialist, on lines 4 and 5 of Algorithm~\ref{alg:specialist}: once on the prototype winner with budget $N{=}20$, once on the runner-up with budget $N{=}6$.
The three tiers, \textsc{Exploring}, \textsc{Optimizing}, and \textsc{FineTune}, progress under auto-escalation.
Two consecutive non-improvements advance the tier; stagnation at \textsc{FineTune} exits the loop; revert-on-regression keeps a bad change from carrying into the next step.

\begin{algorithm}[H]
\caption{\textsc{AdaptiveRefine}: linear refinement loop with auto-escalation and revert-on-regression.}
\label{alg:adaptive_refine}
\begin{algorithmic}[1]
\Procedure{AdaptiveRefine}{$m$, $\Lambda$, $N$}
    \State $\textit{tier} \gets \textsc{Exploring}$;\ \ $\textit{best} \gets m$;\ \ $c \gets 0$
    \For{$i = 1$ to $N$}
        \State $p \gets \textsc{LLMProposal}(\textit{best}, \textit{tier}, \text{history}, \Lambda)$
        \State $s \gets \textsc{Run}(p)$
        \If{$s$ improves on $\text{score}(\textit{best})$}
            \State $\textit{best} \gets p$;\ \ $c \gets 0$
        \Else
            \State revert $p$;\ \ $c \gets c+1$
        \EndIf
        \If{$c \geq 2$}
            \State \textbf{if} $\textit{tier} = \textsc{FineTune}$ \textbf{then} \textbf{break} \hfill$\triangleright$ stagnation exit
            \State \textbf{else} advance $\textit{tier}$;\ \ $c \gets 0$ \hfill$\triangleright$ auto-escalate
            \State \textbf{end if}
        \EndIf
    \EndFor
    \State \Return $\textit{best}$
\EndProcedure
\end{algorithmic}
\end{algorithm}

\needspace{30\baselineskip}
\section{Prompts}
\label{app:prompts}

We include the four key prompts that drive the system.

\needspace{30\baselineskip}
\subsection{Prototype Screen}
The prototype screen prompt (Figure~\ref{fig:prompt_phase0}) is used at the start of each competition. It injects the resource probe and any accumulated skills, and asks the LLM for three diverse model proposals.
\input{fig_prompt_phase0}

\needspace{30\baselineskip}
\subsection{Refinement Proposal}
The refinement proposal prompt (Figure~\ref{fig:prompt_refinement}) is the per-iteration call inside AdaptiveRefine. It carries the current tier, compressed history, and the loaded skills, and uses the six-mode failure taxonomy to structure the next change.
\input{fig_prompt_refinement}

\needspace{30\baselineskip}
\subsection{Learning Production}
The learning production prompt (Figure~\ref{fig:prompt_learnings}) closes each competition. The specialist reflects on its full experiment history and emits two to five plain-text learnings, each with a proposed tier.
\input{fig_prompt_learnings}

\needspace{30\baselineskip}
\subsection{Skill Promotion}
The skill promotion prompt (Figure~\ref{fig:prompt_promotion}) runs between rounds. The orchestrator evaluates each new learning against existing skills and decides \textsc{skip}, \textsc{competition}, \textsc{domain}, \textsc{global}, or \textsc{conflict}.
\input{fig_prompt_promotion}

\needspace{30\baselineskip}
\section{Per-Competition Main Benchmark Results}
\label{app:main_results}

Table~\ref{tab:main_results} reports per-competition results on MLE-Bench Lite.

\input{tab_main_results}

\needspace{30\baselineskip}
\section{Cold Run vs.\ Warm Run}
\label{app:rerun}

Table~\ref{tab:rerun} compares cold-run and warm-run outcomes across all 22 MLE-Bench Lite competitions.
Nine competitions already medaled on the cold run (with few or no accumulated skills) and were skipped in the rerun phase.
Of the 13 that failed the cold run, 8 flipped to medal on the warm run with accumulated global and domain skills, accounting for the lift from 40.9\% to 77.3\% medal rate.
The remaining 5 failed both attempts.

\input{tab_rerun_experiment}

\needspace{30\baselineskip}
\section{Representative Skills}
\label{app:skill_examples}

Table~\ref{tab:skill_examples} shows representative entries from the skill hierarchy with their source competition and an estimate of how many refinement iterations they saved on later, similar tasks.
The full skill inventory contains 5 global, 15 vision, 12 NLP, 19 tabular, and 108 competition entries.

\input{tab_skill_examples}

\needspace{20\baselineskip}
\section{Ablation Resource Breakdown}
\label{app:resources}

Table~\ref{tab:ablation_resources} reports the full resource usage of the three ablation conditions: medal rate, mean test score, output token consumption, wall time, experiments attempted (with success rate), and skills loaded.
The headline finding (flat doubles empty's token cost for fewer medals) is in \S\ref{sec:ablation}; this table provides the supporting detail.

\input{tab_ablation_resources}

\needspace{18\baselineskip}
\section{Ablation Per-Competition Table}
\label{app:ablation_table}

Table~\ref{tab:ablation_results} gives the full per-competition scores and medals for the controlled ablation (Figure~\ref{fig:ablation} in the main text).

\input{tab_ablation_results}

\end{document}

%% file: tab_related_work.tex
\begin{table}[H]
\centering
\small
\caption{Cross-task knowledge mechanisms in ML engineering agents. \checkmark{} = present, \ding{55}{} = absent, $\sim$ = partial.}
\label{tab:related_work}
\setlength{\tabcolsep}{4pt}
\resizebox{\columnwidth}{!}{%
\begin{tabular}{llcccc}
\toprule
\textbf{System} & \textbf{Organization} & \textbf{Cross-Task} & \textbf{Hierarchy} & \textbf{Promotion} & \textbf{MLE-Bench} \\
\midrule
AIDE & None & \ding{55} & \ding{55} & \ding{55} & \checkmark \\
MLE-STAR & Flat (external) & \ding{55} & \ding{55} & \ding{55} & \checkmark \\
R\&D-Agent & Flat (per-task) & $\sim$ & \ding{55} & \ding{55} & \checkmark \\
PiEvolve\tablefootnote{\url{https://github.com/FractalAIResearchLabs/PiEvolve}} & Flat (per-task) & \ding{55} & \ding{55} & \ding{55} & \checkmark \\
MLEvolve & Flat (per-task) & $\sim$ & \ding{55} & \ding{55} & \checkmark \\
MARS & Flat (cross-branch) & $\sim$ & \ding{55} & \ding{55} & \checkmark \\
Agent K v1.0 & Flat (shared) & \checkmark & \ding{55} & \ding{55} & \ding{55} \\
MLCopilot & Flat (vector DB) & \checkmark & \ding{55} & \ding{55} & \ding{55} \\
ExpeL & Flat (vector store) & \checkmark & \ding{55} & \ding{55} & \ding{55} \\
MLZero & Semantic + Episodic & \checkmark & \ding{55} & \ding{55} & $\sim$ \\
Voyager & Flat (embedding) & \checkmark & $\sim$ & \ding{55} & \ding{55} \\
ADAS & Flat (archive) & \checkmark & \ding{55} & \ding{55} & \ding{55} \\
\midrule
\rowcolor{blue!8}
\textbf{HASTE (ours)} & \textbf{3-tier hierarchy} & \checkmark & \checkmark & \checkmark & \checkmark \\
\bottomrule
\end{tabular}
}
\end{table}

%% file: fig_architecture.tex
\begin{figure*}[t]
\centering
\begin{tikzpicture}[
    arrow/.style={->, >=stealth, thick},
    sarrow/.style={->, >=stealth, semithick},
]

\node[draw, rounded corners, fill=blue!10, minimum width=7cm, minimum height=0.7cm, align=center, font=\small] (orch) at (0, 0)
    {\textbf{Orchestrator} {\scriptsize~--- classify, schedule, promote}};

\draw[thick, rounded corners, fill=green!4] (-5.5, -1.4) rectangle (5.5, -5.0);
\node[font=\small\bfseries, anchor=north west] at (-5.3, -1.5) {Specialist {\normalfont\scriptsize(one per domain: Tabular / NLP / Vision)}};

\node[draw, rounded corners=3pt, fill=white, minimum width=1.8cm, minimum height=0.7cm, align=center, font=\scriptsize] (c1) at (-3.8, -2.8) {Task\\[-1pt]Profiler};
\node[draw, rounded corners=3pt, fill=white, minimum width=1.8cm, minimum height=0.7cm, align=center, font=\scriptsize] (p0) at (-1.5, -2.8) {Prototype\\[-1pt]Screen};
\node[draw, rounded corners=3pt, fill=white, minimum width=1.8cm, minimum height=0.7cm, align=center, font=\scriptsize] (refine) at (1.1, -2.8) {Adaptive\\[-1pt]Refinement};
\node[draw, rounded corners=3pt, fill=white, minimum width=1.8cm, minimum height=0.7cm, align=center, font=\scriptsize] (ens) at (3.6, -2.8) {Ensemble};

\draw[sarrow] (c1) -- (p0);
\draw[sarrow] (p0) -- (refine);
\draw[sarrow] (refine) -- (ens);

\node[draw, rounded corners=3pt, fill=orange!12, minimum width=2.2cm, minimum height=0.6cm, align=center, font=\scriptsize] (learn) at (3.6, -4.2) {Produce Learnings};
\draw[sarrow] (ens.south) -- (learn.north);

\node[font=\tiny, text=gray!70, align=center] at (-3.8, -3.45) {metadata,\\CV strategy};
\node[font=\tiny, text=gray!70, align=center] at (-1.5, -3.45) {3 diverse\\models};
\node[font=\tiny, text=gray!70, align=center] at (1.1, -3.45) {explore $\to$ optimize\\$\to$ fine-tune};
\node[font=\tiny, text=gray!70, align=center] at (3.6, -3.45) {rank-average\\top-3};

\draw[thick, rounded corners] (-5.5, -5.8) rectangle (5.5, -8.8);
\node[font=\small\bfseries, anchor=north west] at (-5.3, -5.9) {3-Tier Skill Hierarchy};

\node[draw, dashed, rounded corners, fill=yellow!12, minimum width=10cm, minimum height=0.5cm, font=\scriptsize] (global) at (0, -6.7) {\textbf{Global} (5 entries) --- loaded by all specialists};
\node[draw, dashed, rounded corners, fill=orange!8, minimum width=10cm, minimum height=0.5cm, font=\scriptsize] (dom) at (0, -7.5) {\textbf{Domain} --- Tabular (19) $|$ NLP (12) $|$ Vision (15) --- matching specialist only};
\node[draw, dashed, rounded corners, fill=red!6, minimum width=10cm, minimum height=0.5cm, font=\scriptsize] (comp) at (0, -8.3) {\textbf{Competition} (21 dirs) --- per-task};


\draw[arrow] (0, -0.35) -- node[right, font=\scriptsize] {assign competitions} (0, -1.4);

\draw[arrow, blue!60] (-3.5, -5.8) -- node[left, font=\scriptsize] {load relevant skills} (-3.5, -5.0);

\draw[arrow, orange!60] (3.6, -5.0) -- node[right, font=\scriptsize] {save learnings} (3.6, -5.8);

\draw[arrow, red!60, thick, dashed]
    (5.5, -7.5) -- (6.2, -7.5) -- (6.2, 0) -- node[above, font=\scriptsize, pos=0.9] {\textit{promote between rounds}} (5.1, 0);

\end{tikzpicture}
\caption{HASTE architecture. The Orchestrator assigns competitions to domain Specialists. Each Specialist loads relevant skills, executes the pipeline (profile $\to$ prototype $\to$ refine $\to$ ensemble), and produces learnings. Between rounds, the Orchestrator promotes generalizable learnings upward through the hierarchy via LLM-driven abstraction.}
\label{fig:architecture}
\end{figure*}

%% file: tab_leaderboard.tex
\begin{table}[H]
\centering
\small
\caption{MLE-Bench Lite reference points. Reference numbers from the public MLE-Bench leaderboard (\url{https://github.com/openai/mle-bench}, accessed June 2026).}
\label{tab:leaderboard}
\setlength{\tabcolsep}{6pt}
\begin{tabular}{lllc}
\toprule
\textbf{Agent} & \textbf{LLM} & \textbf{Medal \%} & \textbf{Time} \\
\midrule
Famou-Agent 2.0 & Gemini-3-Pro & 80.3 & 24h \\
MLEvolve & Gemini-3-Pro & 80.3 & 12h \\
PiEvolve & Gemini-3-Pro & 80.3 & 24h \\
CAIR MARS+ & Gemini-3-Pro & 78.8 & 24h \\
\rowcolor{blue!8}
\textbf{HASTE (ours)} & \textbf{Claude Sonnet 4.6} & \textbf{77.3} & \textbf{12h} \\
AIBuildAI & Claude Opus 4.6 & 77.3 & 24h \\
Famou-Agent 2.0 & Gemini-2.5-Pro & 75.8 & 24h \\
ML-Master 2.0 & Deepseek-V3.2 & 75.8 & 24h \\
CAIR MARS & Gemini-3-Pro & 74.2 & 24h \\
PiEvolve (12h) & Gemini-3-Pro & 74.2 & 12h \\
\bottomrule
\end{tabular}
\end{table}

%% file: fig_ablation.tex
\begin{figure}[!tbp]
\centering
\includegraphics[width=\columnwidth]{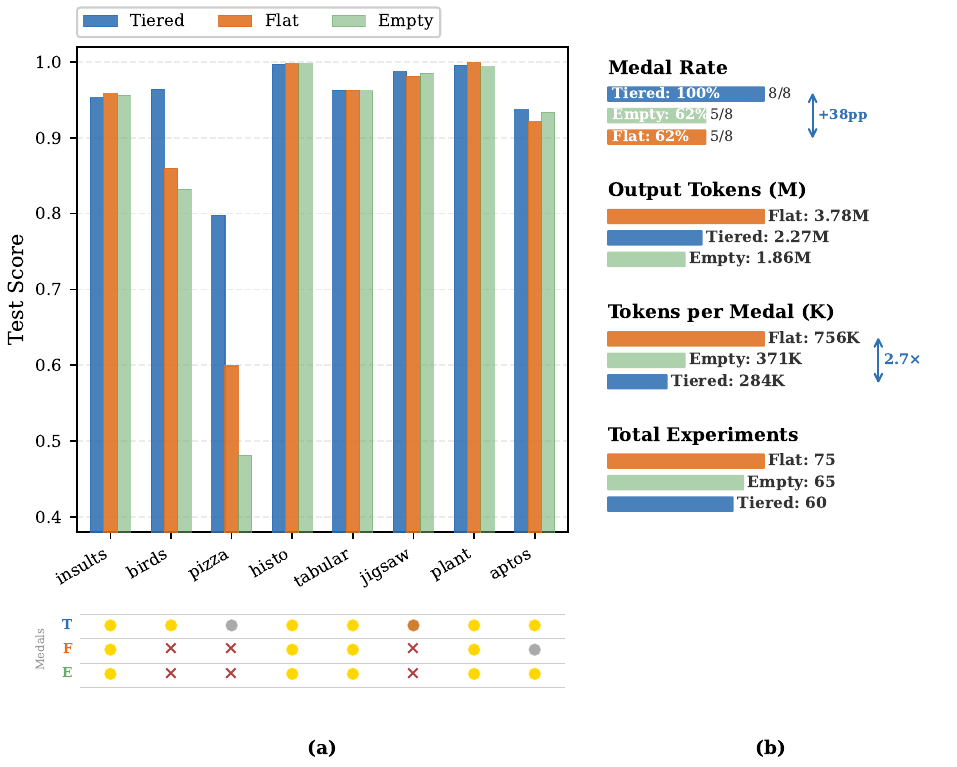}
\caption{\textbf{Controlled ablation in Section~\ref{sec:ablation}.} All conditions use the same 159-skill inventory, model, pipeline, and budget. Tiered loading medals on all 8 competitions; flat and empty both medal on 5 of 8.}
\label{fig:ablation}
\end{figure}

%% file: fig_prompt_phase0.tex
\begin{figure}[H]
\begin{promptbox}{Prototype Screen Prompt (abbreviated)}
\small\ttfamily
You are an ML engineer starting a Kaggle competition.\\[2pt]
Competition: \{competition\_id\}\\
Domain: \{domain\} \quad Metric: \{metric\} (\{direction\})\\
Data: \{rows\} rows, \{features\} features\\
Target: \{target\_type\}\\[2pt]
\textrm{\textit{[Resource budget: GPU model, VRAM, CPU cores, RAM]}}\\
\textrm{\textit{[Accumulated skills from past competitions, if any]}}\\[4pt]
Phase 0 is a diversity screen across model families --- the winner will be refined later.\\[2pt]
Propose 3 DIVERSE model approaches to try. Each should be a fundamentally different strategy (e.g.\ gradient boosting vs.\ neural net vs.\ linear model, or different preprocessing philosophies).\\[2pt]
Return JSON:\\
\{``models'': [\{``name'': ..., ``description'': ...,\\
\quad ``change\_specification'': ...\}, ...]\}
\end{promptbox}
\caption{Abbreviated prototype screen prompt. The full prompt injects measured GPU/CPU/RAM from a resource probe and up to 2000 characters of accumulated skills. The LLM returns three model specifications, each executed with 1-fold validation on full data.}
\label{fig:prompt_phase0}
\end{figure}

%% file: fig_prompt_refinement.tex
\begin{figure}[H]
\begin{promptbox}{Refinement Proposal Prompt (abbreviated)}
\small\ttfamily
You are an ML engineer working on a Kaggle competition.\\[2pt]
Competition: \{competition\_id\}\\
Metric: \{metric\} (\{direction\})\\
Data: \{rows\} rows, \{features\} features\\
Current best score: \{best\_score\}\\
Phase: \{tier\_label\}\\
\{tier\_guidance\}\\[2pt]
\textrm{\textit{[Resource budget, experiment history, accumulated skills,}}\\
\textrm{\textit{current code (first 3000 chars)]}}\\[4pt]
\#\# Failure Mode Diagnosis\\
First, diagnose which failure mode best describes\\
the current state:\\
1. UNDERFITTING --- train score near baseline, small gap\\
2. OVERFITTING --- large train-val gap, high train score\\
3. FEATURE\_GAP --- top features dominate, score plateaus\\
4. NOISE\_CEILING --- high CV variance, score fluctuates\\
5. DISTRIBUTION\_MISMATCH --- train-val-test disagreement\\
6. DIMINISHING\_RETURNS --- each iter improves <0.1\%\\[2pt]
Mention which mode applies and which past skill (if any) influenced your decision.\\[2pt]
Propose ONE atomic change to improve the score.\\
Return JSON: \{``plan'': ..., ``change\_specification'': ...,\\ ``decision'': ``CONTINUE'' | ``NEXT\_TIER'' | ``STOP''\}
\end{promptbox}
\caption{Abbreviated refinement proposal prompt. The tier label rotates through Exploring, Optimizing, and Fine-tuning. The six-mode failure taxonomy gives the LLM structured diagnostic guidance. The \texttt{decision} field allows the LLM to self-escalate tiers or terminate early.}
\label{fig:prompt_refinement}
\end{figure}

%% file: fig_prompt_learnings.tex
\begin{figure}[H]
\begin{promptbox}{Learning Production Prompt (abbreviated)}
\small\ttfamily
You just finished working on ``\{competition\_id\}''.\\
Domain: \{domain\} \quad Metric: \{metric\} (\{direction\})\\
Dataset: \{rows\} rows, \{features\} features\\[2pt]
\#\# Your Experiments and Results\\
\textrm{\textit{[Full experiment history with scores, deltas, kept/reverted status]}}\\[4pt]
Write learnings that could help in future competitions.\\[2pt]
Distinguish SUCCESS and FAILURE learnings:\\
- What worked --- ``this technique improved the score because...''\\
- What failed --- ``this technique hurt the score because...''\\[2pt]
Failure learnings are equally valuable. Knowing what NOT to do saves future compute.\\[2pt]
For failure learnings, note:\\
- WHY it failed (overfitting? wrong model family?)\\
- WHEN it might fail again (conditions to watch for)\\[2pt]
Classify each learning:\\
- ``competition'' --- specific to this task only\\
- ``domain'' --- relevant to similar \{domain\} tasks\\
- ``global'' --- relevant to any ML task\\[2pt]
Return JSON: \{``learnings'': [\{``title'': ...,\\
\quad ``body'': ..., ``proposed\_tier'': ...\}, ...]\}\\
Write 2--5 learnings. Focus on actionable, non-obvious insights.
\end{promptbox}
\caption{The learning production prompt. Each specialist reflects on its full experiment history after completing a competition. Learnings are saved to the competition tier and later evaluated for promotion by the orchestrator.}
\label{fig:prompt_learnings}
\end{figure}

%% file: fig_prompt_promotion.tex
\begin{figure}[H]
\begin{promptbox}{Skill Promotion Prompt (abbreviated)}
\small\ttfamily
You are reviewing ML learnings produced by domain agents after completing competitions.
Your job is to decide which learnings should be promoted up the skill hierarchy.\\[2pt]
\textrm{\textit{[All existing global and domain skills, for deduplication]}}\\
\textrm{\textit{[New learnings from all domain agents this round]}}\\[4pt]
\#\# Promotion Rules\\
For each learning, decide:\\
1. ``skip'' --- already covered or too obvious\\
2. ``competition'' --- too specific (dataset quirks, row indices)\\
3. ``domain'' --- generalizable to similar tasks. Abstract it.\\
4. ``global'' --- universally useful across all ML tasks\\
5. ``conflict'' --- contradicts an existing learning. Note conditions under which each holds.\\[4pt]
\#\# Quality Standards\\
- Be selective. Promote AT MOST 50\% of learnings.\\
- Abstractions MUST NOT mention specific competition\\
\quad names, dataset names, or exact score values.\\
\quad Bad: ``On aerial-cactus, AUC reached 0.9997''\\
\quad Good: ``When AUC is near ceiling (>0.999),\\
\quad\quad further refinement yields diminishing returns''\\
- For ``conflict'': provide both the conflicting skill\\
\quad ID and a condition annotation.
\end{promptbox}
\caption{The skill promotion prompt. The orchestrator evaluates all new learnings against existing skills after each round. Abstractions strip competition-specific details to produce reusable domain or global knowledge.}
\label{fig:prompt_promotion}
\end{figure}

%% file: tab_main_results.tex
\begin{table}[H]
\centering
\caption{Per-competition results on MLE-Bench Lite (22 competitions). --- indicates the value was not recorded for that run.}
\label{tab:main_results}
\resizebox{\textwidth}{!}{%
\begin{tabular}{llcrcccc}
\toprule
\textbf{Competition} & \textbf{Domain} & \textbf{Medal} & \textbf{Test Score} & \textbf{Phase0} & \textbf{Final CV} & \textbf{Hours} & \textbf{Iters} \\
\midrule
aerial cactus identification & Vision & \textbf{Gold} & 1.00000 & 0.9994 & 1.0000 & 1.8 & 5 \\
aptos2019 blindness detection & Vision & \textbf{Gold} & 0.93723 & 0.9178 & 0.9280 & 11.5 & 1 \\
denoising dirty documents & Vision & \textbf{Gold} & 0.01127 & 0.0796 & 0.0122 & 10.5 & 6 \\
histopathologic cancer detection & Vision & \textbf{Gold} & 0.99692 & --- & --- & 12.0 & --- \\
plant pathology 2020 fgvc7 & Vision & \textbf{Gold} & 0.99637 & --- & --- & 12.0 & --- \\
the icml 2013 whale challenge ri... & Vision & \textbf{Gold} & 0.99364 & --- & --- & 12.0 & --- \\
dogs vs cats redux kernels edition & Vision & Bronze & 0.05514 & --- & --- & 12.0 & --- \\
siim isic melanoma classification & Vision & Bronze & 0.93730 & 0.8890 & 0.9236 & 11.6 & 2 \\
dog breed identification & Vision & --- & 0.58682 & --- & --- & 12.0 & --- \\
ranzcr clip catheter line classi... & Vision & --- & 0.95658 & --- & --- & 12.0 & --- \\
detecting insults in social comm... & NLP & \textbf{Gold} & 0.95319 & 0.9026 & 0.9525 & 9.8 & 18 \\
random acts of pizza & NLP & Silver & 0.79847 & 0.7648 & 0.7916 & 1.8 & 13 \\
spooky author identification & NLP & Silver & 0.22115 & 0.3553 & 0.3040 & 12.0 & 11 \\
jigsaw toxic comment classificat... & NLP & Bronze & 0.98653 & 0.9621 & 0.9925 & 12.0 & 4 \\
text normalization challenge eng... & NLP & Bronze & 0.99038 & --- & --- & 12.0 & --- \\
text normalization challenge rus... & NLP & Bronze & 0.97888 & --- & 0.9791 & 10.5 & 10 \\
nomad2018 predict transparent co... & Tabular & \textbf{Gold} & 0.05470 & 0.0619 & 0.0533 & 6.6 & 11 \\
tabular playground series dec 2021 & Tabular & \textbf{Gold} & 0.96300 & 0.9616 & 0.9621 & 10.6 & 1 \\
leaf classification & Tabular & --- & 0.06652 & 0.5442 & 0.0498 & 2.9 & 11 \\
new york city taxi fare prediction & Tabular & --- & 6.54783 & 3.5673 & 3.2401 & 11.8 & 8 \\
tabular playground series may 2022 & Tabular & --- & 0.99630 & 0.9461 & 0.9959 & 10.9 & 11 \\
mlsp 2013 birds & Audio & \textbf{Gold} & 0.96446 & 0.8094 & 0.9198 & 11.2 & 11 \\

\bottomrule
\end{tabular}%
}
\end{table}

%% file: tab_rerun_experiment.tex
\begin{table}[H]
\centering
\caption{Cold run vs.\ warm run across all 22 MLE-Bench Lite competitions. Cold = first attempt (few or no accumulated skills); Warm = later attempt with accumulated skills. Nine competitions already medaled cold and were not re-attempted. Of the 13 that failed cold, 8 flipped to medal on the warm run.}
\label{tab:rerun}
\small
\setlength{\tabcolsep}{4pt}
\begin{tabular}{@{}lccccl@{}}
\toprule
\textbf{Competition} & \textbf{Cold} & \textbf{Cold} & \textbf{Warm} & \textbf{Warm} & \\
 & \textbf{Skills} & \textbf{Medal} & \textbf{Skills} & \textbf{Medal} & \textbf{Outcome} \\
\midrule
\multicolumn{6}{l}{\textit{Already medaled on cold run (not re-attempted)}} \\
\addlinespace
denoising-dirty-documents & 5 & \textbf{G} & --- & --- & already medaled \\
detecting-insults & 5 & \textbf{G} & --- & --- & already medaled \\
dogs-vs-cats & 37 & B & --- & --- & already medaled \\
histopathologic & 14 & \textbf{G} & --- & --- & already medaled \\
plant-pathology & 14 & \textbf{G} & --- & --- & already medaled \\
random-acts-of-pizza & 5 & S & --- & --- & already medaled \\
spooky-author & 5 & S & --- & --- & already medaled \\
tabular-playground-dec & 21 & \textbf{G} & --- & --- & already medaled \\
whale-challenge & 14 & \textbf{G} & --- & --- & already medaled \\
\midrule
\multicolumn{6}{l}{\textit{Failed cold, flipped to medal on warm run}} \\
\addlinespace
aerial-cactus & 5 & --- & 50 & \textbf{G} & flipped \\
aptos2019 & 34 & --- & 55 & \textbf{G} & flipped \\
jigsaw-toxic & 37 & --- & 59 & B & flipped \\
mlsp-2013-birds & 32 & --- & 60 & \textbf{G} & flipped \\
nomad2018 & 14 & --- & 60 & \textbf{G} & flipped \\
siim-isic-melanoma & 50 & --- & 71 & B & flipped \\
text-norm-english & 50 & --- & 65 & B & flipped \\
text-norm-russian & 50 & --- & 72 & B & flipped \\
\midrule
\multicolumn{6}{l}{\textit{Failed both cold and warm runs}} \\
\addlinespace
dog-breed & 37 & --- & 62 & --- & no medal \\
leaf-classification & 29 & --- & 54 & --- & no medal \\
nyc-taxi-fare & 59 & --- & 68 & --- & no medal \\
ranzcr-catheter & 37 & --- & 60 & --- & no medal \\
tabular-playground-may & 21 & --- & 60 & --- & no medal \\
\bottomrule
\end{tabular}
\end{table}

%% file: tab_skill_examples.tex
\begin{table}[H]
\centering
\caption{Representative skills from the 3-tier hierarchy. Each entry is a plain-text file traced to the experiment that produced it. Iteration savings are estimated from cold-start vs.\ skill-loaded runs of the same competition.}
\label{tab:skill_examples}
\small
\begin{tabular}{@{}llp{4.3cm}lc@{}}
\toprule
\textbf{Tier} & \textbf{Type} & \textbf{Skill (abbreviated)} & \textbf{Source} & \textbf{Iters saved} \\
\midrule
Global & Tech. & Ensembling a strong model with a weaker one can degrade performance & dogs-vs-cats & $\sim$2 each \\
\addlinespace
Global & Tech. & Larger architecture does not fix wrong problem formulation (ordinal vs.\ classification) & aptos2019 & $\sim$1--2 \\
\addlinespace
Global & Tech. & Chance-level scores usually indicate a training bug, not a model limitation & detecting-insults & $\sim$1--2 \\
\midrule
Domain\,(NLP) & Tech. & DeBERTa-v3-base is the strongest starting point for text classification under 12h & detecting-insults & $\sim$2--3 \\
\addlinespace
Domain\,(Vis.) & Tech. & ConvNeXt-Large + 10-fold stratified CV is a strong default for small-dataset image classification & plant-pathology & $\sim$1--2 \\
\addlinespace
Domain\,(Tab.) & Tech. & Log-transforming the target can hurt RMSE on right-skewed regression targets & NYC-taxi-fare & $\sim$1 \\
\midrule
Domain\,(Vis.) & Refine & Vision refinement hints: 3/3 changes kept (100\% hit rate). Try first: switch to ConvNeXt, add TTA & aerial-cactus & guided \\
\addlinespace
Domain\,(NLP) & Refine & NLP refinement hints: 5/12 changes kept (42\%). Try first: fix LR, add FGM adversarial training & detecting-insults & guided \\
\midrule
Comp. & Tech. & Crystal descriptors (SOAP, Ewald, PRDF) cause regression on small crystal datasets (<5K samples) & nomad2018 & $\sim$2 \\
\bottomrule
\end{tabular}
\end{table}

%% file: tab_ablation_resources.tex
\begin{table}[H]
\centering
\caption{Resource usage under three skill-loading conditions. Flat uses $2\times$ the output tokens of Empty without improving medal rate.}
\label{tab:ablation_resources}
\begin{tabular}{lccc}
\toprule
\textbf{Metric} & \textbf{Tiered} & \textbf{Flat} & \textbf{Empty} \\
\midrule
Medal rate & \textbf{100\%} (8/8) & 62.5\% (5/8) & 62.5\% (5/8) \\
Mean test score & \textbf{0.949} & 0.910 & 0.893 \\
\midrule
Total output tokens & 2.27M & 3.78M & \textbf{1.86M} \\
Mean output tokens & 284K & 472K & \textbf{232K} \\
Tokens per medal & \textbf{284K} & 756K & 372K \\
\midrule
Mean wall time (h) & \textbf{10.2} & 10.5 & 10.7 \\
Total experiments & 60 & 75 & 65 \\
\midrule
Skills loaded (mean) & 36.9 & 159 & 0 \\
Context from skills & $\sim$25K chars & $\sim$145K chars & 0 \\
\bottomrule
\end{tabular}
\end{table}

%% file: tab_ablation_results.tex
\begin{table}[H]
\centering
\small
\caption{Controlled ablation: scores and medals (G/S/B/---) under three skill-loading conditions (8 competitions, 1 seed). \textbf{Tiered}: domain-scoped; \textbf{Flat}: all 159 skills; \textbf{Empty}: none. Best score per competition in \textbf{bold}.}
\label{tab:ablation_results}
\setlength{\tabcolsep}{5pt}
\begin{tabular}{llccc}
\toprule
\textbf{Competition} & \textbf{Domain} & \textbf{Tiered} & \textbf{Flat} & \textbf{Empty} \\
\midrule
detecting-insults & NLP & 0.953 G & \textbf{0.958} G & 0.956 G \\
jigsaw-toxic & NLP & \textbf{0.987} B & 0.981 --- & 0.985 --- \\
random-acts-of-pizza & NLP & \textbf{0.798} S & 0.599 --- & 0.481 --- \\
histopathologic & Vision & 0.997 G & \textbf{0.998} G & \textbf{0.998} G \\
plant-pathology & Vision & 0.996 G & \textbf{0.999} G & 0.994 G \\
aptos2019 & Vision & \textbf{0.937} G & 0.921 S & 0.934 G \\
tabular-playground & Tabular & \textbf{0.963} G & \textbf{0.963} G & 0.962 G \\
mlsp-2013-birds & Audio & \textbf{0.964} G & 0.860 --- & 0.832 --- \\
\midrule
\textbf{Medal rate} & & \textbf{100\% (8/8)} & 62.5\% (5/8) & 62.5\% (5/8) \\
\textbf{Mean score} & & \textbf{0.949} & 0.910 & 0.893 \\
\bottomrule
\end{tabular}
\end{table}